# Multi-Relation Aware Temporal Interaction Network Embedding

Ling Chen*

College of Computer Science and Technology, Zhejiang University, Hangzhou 310027, China, lingchen@cs.zju.edu.cn

Shanshan Yu

College of Computer Science and Technology, Zhejiang University, Hangzhou 310027, China, shshyu@cs.zju.edu.cn

Dandan Lyu

College of Computer Science and Technology, Zhejiang University, Hangzhou 310027, China, revaludo@cs.zju.edu.cn

Da Wang

College of Computer Science and Technology, Zhejiang University, Hangzhou 310027, China, wangda9655@cs.zju.edu.cn

Temporal interaction networks are formed in many fields, e.g., e-commerce, online education, and social network service. Temporal interaction network embedding can effectively mine the information in temporal interaction networks, which is of great significance to the above fields. Usually, the occurrence of an interaction affects not only the nodes directly involved in the interaction (interacting nodes), but also the neighbor nodes of interacting nodes. However, existing temporal interaction network embedding methods only use historical interaction relations to mine neighbor nodes, ignoring other relation types. In this paper, we propose a multi-relation aware temporal interaction network embedding method (MRATE). Based on historical interactions, MRATE mines historical interaction relations, common interaction relations, and interaction sequence similarity relations to obtain the neighbor based embeddings of interacting nodes. The hierarchical multi-relation aware aggregation method in MRATE first employs graph attention networks (GATs) to aggregate the interaction impacts propagated through a same relation type and then combines the aggregated interaction impacts from multiple relation types through the self-attention mechanism. Experiments are conducted on three public temporal interaction network datasets, and the experimental results show the effectiveness of MRATE.

## 1 INTRODUCTION

Users interact with different items at different time in many fields, e.g., e-commerce (customers purchase products) [1], online education (students participate in MOOC courses) [2], and social network service (users post in Reddit groups) [3, 4]. These interactions between users and items form a temporal interaction network, which pays more attention to interaction time compared to a static interaction network. Figure 1 shows an example of a temporal interaction network consisting of 3 users and 4 items. Each edge in a temporal interaction network is marked with a time stamp, which indicates the time when the user and the item interacted with each other. Temporal interaction network embedding aims at learning the low-dimensional representations of users and items that can effectively capture their dynamics. The effective representations of users and items are of great significance for a variety of fundamental tasks [5-10], e.g., product recommendation in e-commerce, course recommendation in online education, and group recommendation in social network service.

---

* Corresponding author. Tel: +86 13606527774.

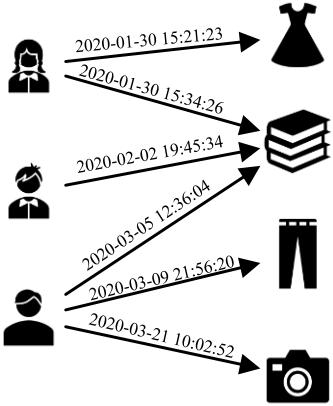

Figure 1: An example of a temporal interaction network.

Existing temporal interaction network embedding methods include non-graph structure based embedding methods and graph structure based embedding methods. The former use non-graph structures (e.g., matrices or sequences) to represent user-item interactions. Early studies introduce time slots into a user-item matrix and then utilize traditional latent factor models to embed users and items [11-13], which ignore the order of user-item interactions when forming matrices. With the development of recurrent neural networks (RNNs), the sequence information of temporal interaction networks is also exploited [14-17]. Most of these methods consider dynamic properties only for users, e.g., taking static item embeddings as inputs to update user embeddings, which ignore the dynamics of items.

Utilizing graph structures to represent user-item interactions is a promising way to mine richer information based on historical interactions. Early graph structure based embedding methods usually take time slots as nodes in a graph [18, 19], which essentially use a static graph and cannot effectively model the dynamics of users and items. Recently, some researchers directly embed nodes (users and items) in a temporal interaction network into a shared low-dimensional vector space [20-25]. In addition to updating the embeddings of interacting nodes at every interaction, these methods also consider the interaction impacts propagated by neighbor nodes. However, these methods only use historical interaction relations, ignoring other relation types.

To address the aforementioned problems, we propose a multi-relation aware temporal interaction network embedding method (MRATE). Based on historical interactions, MRATE mines the relations of multiple types between nodes to build local relation graphs for interacting nodes, which are then used for aggregating information from neighbors.

The main contributions of this paper are as follows:

1) Propose MRATE, which mines not only historical interaction relations, but also common interaction relations and interaction sequence similarity relations, from historical interactions to obtain the neighbor based embeddings of interacting nodes.

2) Introduce a hierarchical multi-relation aware aggregation method, which first employs graph attention networks (GATs) to aggregate the interaction impacts propagated through a same relation type and then combines the aggregated interaction impacts from multiple relation types through the self-attention mechanism.

3) Evaluate MRATE on three public temporal interaction network datasets and make comparisons with the state-of-the-art temporal interaction network embedding methods. Experimental results show that MRATE can achieve superior performance.

The rest of this paper is organized as follows. Section 2 reviews the related work. Section 3 gives the preliminaries of this paper and defines the research problem. Section 4 introduces the proposed method MRATE in detail. Section 5 presents the experimental settings and results. Finally, Section 6 concludes the paper and gives a brief discussion of the future work.

## 2 RELATED WORK

In this section, we review the previous works related to our work, including non-graph structure based embedding methods and graph structure based embedding methods.

### 2.1 Non-graph Structure Based Embedding Methods

Non-graph Structure based embedding methods use non-graph structures (e.g., matrices or sequences) to represent user-item interactions. In order to model the dynamics of users and items, early studies usually introduce time slots into a user-item matrix and then utilize traditional latent factor models to embed users and items. For instance, Koren et al. [11] extended SVD++ factor model [26] by incorporating time related user and item biases to get the representations of users and items. Yao et al. [12] proposed TenMF, a tensor-based factorization approach that first models the multi-dimensional contextual information of check-in data as a three-order tensor, which represents the relations across users, locations, and time slots, respectively, and then jointly embeds users and locations. Bhargava et al. [13] proposed a framework similar with TenMF, in which a user-location-activity-time tensor is constructed by fusing data from multiple heterogeneous data sources, and tensor factorization is used for embedding. However, these methods ignore the order of user-item interactions when forming matrices.

RNNs and variants, e.g., long short-term memory networks (LSTMs) and gated recurrent unit networks (GRUs), are a kind of machine learning models that specialize in processing sequence data, and have been applied in many applications, e.g., disease recognition [27], human activity recognition [28, 29], and crowd flow prediction [30, 31]. Nowadays, more and more researchers adopt RNNs for temporal network embedding to exploit sequential effects [14-17]. For example, Donkers et al. [14] proposed a novel GRU with additional user input layers to get the representations of users based on historical interactions, which are used to support personalized next item recommendation. Li et al. [15] proposed to obtain user embeddings by integrating both the historical preferences and present motivations, which are all learned from the interactions of users through two LSTMs. Zhu et al. [16] equipped a LSTM with time gates to model time intervals in users' interaction sequences, which can incorporate time information into user embeddings. Wu et al. [17] proposed RRN, a framework that utilizes RNNs to generate dynamic user and item embeddings from rating networks. However, these methods consider dynamic properties only for users, e.g., taking static item embeddings as inputs to update user embeddings, which ignore the dynamics of items.

## 2.2 Graph Structure Based Embedding Methods

Considering the excellent ability of graph structures to describe complex data, researchers begin to utilize graph structures to represent user-item interactions [18-25]. Usually, time slots are regarded as nodes in graphs to model the dynamics of users and items. For example, Xie et al. [18] proposed a graph based embedding model, which captures sequential, geographical, temporal, and semantic effects in a unified way by embedding four corresponding graphs (POI-POI, POI-Region, POI-Time Slot, and POI-Word graphs) into a shared low-dimensional space. Yang et al. [19] first formed a hypergraph including both user-user edges and user-time-POI-semantic hyperedges and then proposed a random walk based method to learn embeddings by the skip-gram model. However, these methods essentially use a static graph and cannot effectively model the dynamics of users and items.

Recently, directly embedding nodes (users and items) in a temporal interaction network into a low-dimensional vector space has attracted more and more attention. These methods can make better use of time information for capturing the dynamic properties of users and items, which supports getting more effective node embeddings. For instance, Nguyen et al. [20] proposed to utilize random walks that obey interaction time to obtain node sequences and then get node embeddings by the skip-gram model. Trivedi et al. [21] described temporal evolution over graphs as association and communication events, and then proposed a two-time scale deep temporal point process approach to update the representations of nodes once they are involved in these events. Kumar et al. [22] proposed JODIE, a framework that adopts two RNNs to update the embeddings of the interacting user and item at every interaction, which can obtain the embedding trajectories of users and items.

Due to the interaction impacts propagated by neighbor nodes, most temporal interaction network embedding methods take neighbor information into consideration. For example, Zuo et al. [23] integrated the Hawkes process into network embedding so as to capture the influence of historical neighbors on current neighbors. Lu et al. [24] proposed to microscopically model the formation process of network structure with a temporal attention point process, which can measure the effects of different neighbor nodes on edge generation, and macroscopically constrain the network structure to obey a certain evolutionary pattern with a dynamic equation. Ma et al. [25] updated the information of a new interaction to the two interacting nodes and also propagated this information to the neighbor nodes through RNNs. Although the above methods have considered neighbor information, they only use historical interaction relations to define neighbor nodes, ignoring other relation types.

## 3 PRELIMINARIES

In this section, we first give the definitions of some basic concepts and terms, as shown in Table 1, and then formally define the problem.

**Definition 1:** (Temporal interaction network) A temporal interaction network is a sequence of interactions sorted by interaction time, and can be denoted as $\mathcal{S} = \{s_i\}_{i=1}^{N}$. An interaction indicating user $u \in \mathcal{U}$ interacts with item $v \in \mathcal{V}$ at time $t \in \mathcal{T}$ can be denoted as $s = (u, v, t)$, where $\mathcal{U}$, $\mathcal{V}$, and $\mathcal{T}$ are the sets of users, items, and interaction time, respectively.

**Definition 2:** (Local relation graph) A local relation graph can be denoted as $\mathcal{G}_{n_i} = (\mathcal{V}_{n_i}, \mathcal{E}_{n_i}, \mathcal{R}_{n_i}, \mathcal{Q}_{n_i})$, where $\mathcal{V}_{n_i}$, $\mathcal{E}_{n_i}$, $\mathcal{R}_{n_i}$, and $\mathcal{Q}_{n_i}$ are the sets of nodes, edges, relation types, and relation attributes related to node $n_i$, respectively. All nodes in local relation graph $\mathcal{G}_{n_i}$ are connected to node $n_i$, and an edge indicating a relation involving another node $n_j$ with relation type $r \in \mathcal{R}_{n_i}$ and relation attribute $q = (t, w) \in \mathcal{Q}_{n_i}$ can be denoted as $e = (n_j, r, q) \in \mathcal{E}_{n_i}$, where $t$ is the time attribute and $w$ is the weight attribute.

Table 1: Notations used in this paper

| Notation | Description |
| --- | --- |
| $\mathcal{U}, \mathcal{V}, \mathcal{T}$ | The sets of users, items, and interaction time. |
| $\mathcal{S}, s$ | Temporal interaction network and interaction. |
| $\boldsymbol{u}_{i,t_{\mathrm{aft}}}, \boldsymbol{v}_{j,t_{\mathrm{aft}}}$ | The embeddings of user $u_i$ and item $v_j$ after the current interaction. |
| $\boldsymbol{u}_{i,t_{\mathrm{prev}}}, \boldsymbol{v}_{j,t_{\mathrm{prev}}}$ | The embeddings of user $u_i$ and item $v_j$ after the previous interactions. |
| $\Delta_{u_i}, \Delta_{v_j}$ | The time intervals between the previous interactions of user $u_i$ and item $v_j$ and the current interaction. |
| $\mathcal{G}_{n_i}$ | The local relation graph of node $n_i$ (a user or an item). |
| $\mathcal{V}_{n_i}, \mathcal{E}_{n_i}, \mathcal{R}_{n_i}, \mathcal{Q}_{n_i}$ | The sets of nodes, edges, relation types, and relation attributes in $\mathcal{G}_{n_i}$. |
| $e, r, q$ | Edge, relation type, and relation attribute. |
| $t, w$ | Time attribute and weight attribute. |
| $\boldsymbol{s}_{n_i}$ | The sequence based embedding of node $n_i$. |
| $\boldsymbol{h}_{n_i}$ | The embedding of node $n_i$ after the previous interaction. |
| $\boldsymbol{h}_{n_i}^{\mathrm{his}}, \boldsymbol{h}_{n_i}^{\mathrm{com}}, \boldsymbol{h}_{n_i}^{\mathrm{seq}}$ | The neighbor based embeddings of node $n_i$ based on historical interaction relations, common interaction relations, and interaction sequence similarity relations. |
| $\boldsymbol{h}'_{n_i}$ | The neighbor based embedding of node $n_i$. |
| $\widetilde{\boldsymbol{v}}_{j,t_{\mathrm{bef}}}$ | The predicted embedding of item $v_j$. |

**Definition 3:** (Current interaction and previous interaction) When a new interaction occurs, this interaction is denoted as the current interaction, and the last interactions of the interacting user and item are denoted as the previous interactions.

Our goal is to learn node embeddings in a temporal interaction network by capturing the dynamics of users and items. We can formally define the problem as follows:

**Definition 4:** (Temporal interaction network embedding) Given a temporal interaction network $\mathcal{S} = \{s_i\}_{i=1}^{N}$, temporal interaction network embedding aims to learn a mapping function $f: \mathcal{U} \cup \mathcal{V} \to \mathbb{R}^d$, where $d$ is the number of embedding dimensions, $d \ll \|\mathcal{U}\|$ and $d \ll \|\mathcal{V}\|$.

## 4 METHODOLOGY

The framework of MRATE is shown in Figure 2. When an interaction occurs, we first mine historical interaction relations, common interaction relations, and interaction sequence similarity relations between nodes from historical interactions to build local relation graphs for interacting nodes. Then, in order to obtain the neighbor based embeddings of interacting nodes, we introduce a hierarchical multi-relation aware aggregation method to aggregate the interaction impacts propagated by related neighbor nodes in corresponding local relation graphs through different relation types. Finally, we employ two RNNs to update the representations of the interacting user and item based on the neighbor based embeddings, the embeddings after their previous interactions, and the time intervals between the previous interactions of interacting nodes and the current interaction.

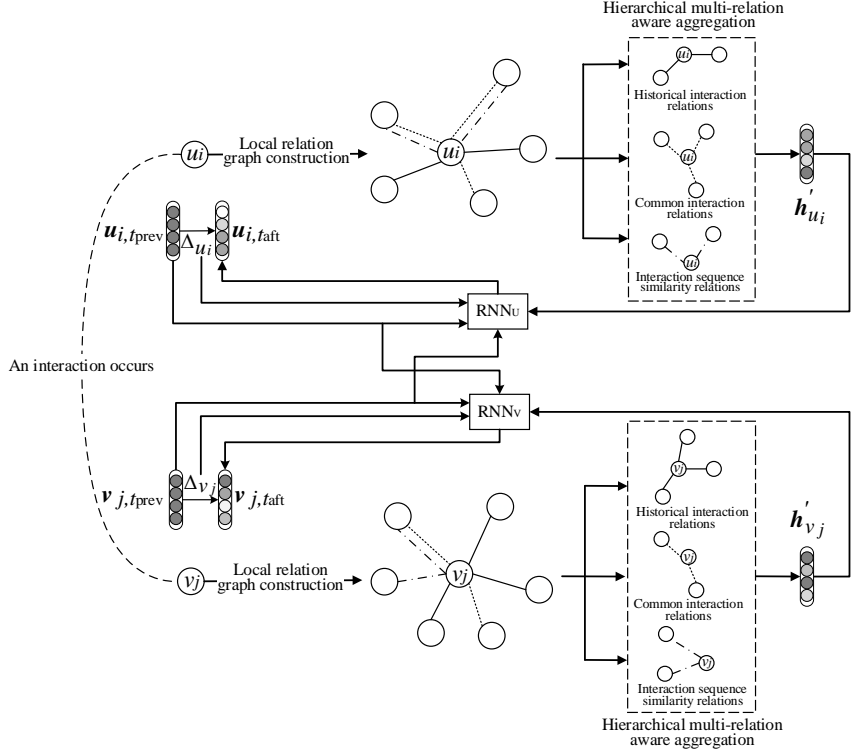

Figure 2: The framework of MRATE.

### 4.1 Local Relation Graph Construction

Considering that the occurrence of an interaction also affects the neighbor nodes of interacting nodes, MRATE mines historical interaction relations, common interaction relations, and interaction sequence similarity relations from historical interactions to capture neighbor information. Here, we first define three kinds of relation types, and then describe the construction of local relation graphs.

*4.1.1 Historical Interaction Relation.*

Due to the phenomenon known as repeat consumption [32], the representations of interacting nodes are influenced not only by the current interaction, but also by historical interactions. Thus, we define historical interaction relations according to whether two nodes have interacted historically. Specifically, if two nodes have interacted in history, there is a historical interaction relation between them. The time attribute $t$ of the historical interaction relation is the last time the two nodes interacted, and the weight attribute $w$ is the number of historical interactions between them.

*4.1.2 Common Interaction Relation.*

Common interaction relations, which are widely used in recommendation systems, e.g., "People who like this movie also like" on Douban and "Customers who buy this product also buy" on Amazon [33], contain significant information for modeling the interaction behaviors of users and items. If two users or two items have interacted with a same node within time slot $T$, there is a common interaction relation between them. Since the time when two nodes interacted with a same node is different, we choose the one closest to the current interaction time as the time attribute $t$ of the common interaction relation. The weight attribute $w$ is the number of historical common interactions between the two nodes.

*4.1.3 Interaction Sequence Similarity Relation.*

It is useful to consider the interaction sequence similarity relation, as it reflects the similarity of the properties of two nodes. We utilize the Doc2Vec model to get the sequence based embeddings of users and items. Doc2Vec [34] is an improvement of Word2Vec [35], which can embed sentences into a low-dimensional vector space and the distances of semantically similar sentences in the space are also close. In this work, the interaction sequences of all users and items are regarded as "documents", and each interaction sequence (i.e., a sequence of users who interact with a same item or a sequence of items which interact with a same user) sorted by interaction time is regarded as a "sentence". Note that, due to the constant occurrence of interactions, the Doc2Vec model is updated by using incremental training to obtain new sequence based embeddings.

Given two nodes of the same type (two users or two items), we calculate the cosine similarity based on their sequence based embeddings, which can be formulated by:
$$cosSim\left(\boldsymbol{s}_{n_i}, \boldsymbol{s}_{n_j}\right) = \boldsymbol{s}_{n_i} \cdot \boldsymbol{s}_{n_j} \qquad (1)$$
where $\cdot$ denotes dot product. $\boldsymbol{s}_{n_i}$ and $\boldsymbol{s}_{n_j}$ are the sequence based embeddings of node $n_i$ and node $n_j$, respectively. If the cosine similarity is larger than the threshold $\mu$, there is an interaction sequence similarity relation between the two nodes. The time attribute $t$ is the last interaction time in the interaction sequences, and the weight attribute $w$ is the cosine similarity.

To reduce computational time, MRATE only constructs the local relation graphs of interacting nodes instead of global relation graphs before an interaction occurs. According to the aforementioned definitions, we can mine nodes that have historical interaction relations, common interaction relations, or interaction sequence similarity relations with interacting nodes, and then construct corresponding local relation graphs.

## 4.2 Hierarchical Multi-Relation Aware Aggregation

Based on the local relation graph of an interacting node, the interaction impacts propagated by neighbor nodes through different relation types may have different impacts on the embedding of the interacting node. Thus, MRATE introduces a hierarchical multi-relation aware aggregation method to obtain the neighbor based embedding of an interacting node, which contains two major components: intra-relation aggregation and inter-relation aggregation, as shown in Figure 3.

Given local relation graph $\mathcal{G}_{n_i}$, the input of hierarchical multi-relation aware aggregation is the interaction impacts propagated by neighbor nodes within the time period between the previous interaction of $n_i$ and the current interaction. For simplicity, here we use the embedding of a node after the previous interaction as the interaction impact propagated by this node. Thus, the input can be defined as $\{\boldsymbol{h}_{n_1}, \boldsymbol{h}_{n_2}, \ldots, \boldsymbol{h}_{n_M}\}$, where $\boldsymbol{h}_{n_j} \in \mathbb{R}^d$ is the embedding of neighbor node $n_j$ after the previous interaction, and $M$ is the number of related neighbor nodes, which are the neighbor nodes that have interacted within the time period.

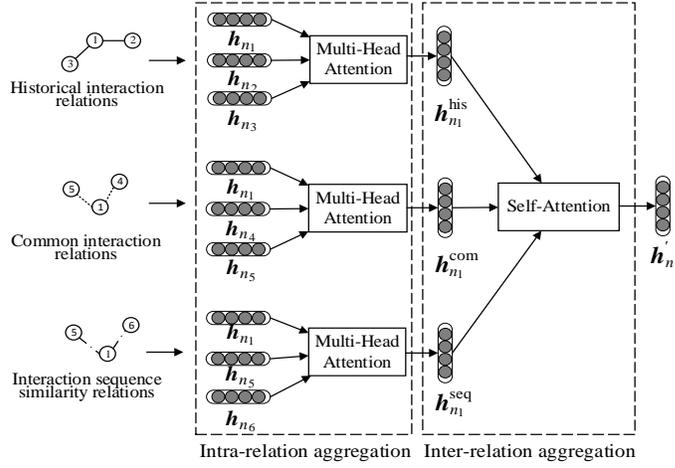

Figure 3: An illustration of hierarchical multi-relation aware aggregation. $\boldsymbol{h}_{n_i}$ denotes the interaction impacts propagated by node $n_i$. $\boldsymbol{h}_{n_i}^{\text{his}}$, $\boldsymbol{h}_{n_i}^{\text{com}}$, and $\boldsymbol{h}_{n_i}^{\text{seq}}$ represent the neighbor based embeddings of $n_i$ based on historical interaction relations, common interaction relations, and interaction sequence similarity relations, respectively. $\boldsymbol{h}'_{n_i}$ is the neighbor based embedding of $n_i$.

### 4.2.1 Intra-Relation Aggregation.

In order to distinguish different relation types, three graph attention layers, which are inspired by GATs [36], are adopted to aggregate the interaction impacts propagated by neighbor nodes through historical interaction relations, common interaction relations, and interaction sequence similarity relations, respectively. The attention mechanism used in these layers is multi-head attention with $K$ heads, which helps to stabilize the learning process. After intra-relation aggregation, we can obtain three embeddings $\boldsymbol{h}_{n_i}^{\text{his}}$, $\boldsymbol{h}_{n_i}^{\text{com}}$, and $\boldsymbol{h}_{n_i}^{\text{seq}}$, which represent the neighbor based embeddings of $n_i$ based on historical interaction relations, common interaction relations, and interaction sequence similarity relations, respectively.

The key idea of intra-relation aggregation is that even if the interaction impacts are propagated through a same relation type, different neighbor nodes may have different significance on the representation of the interacting node. We can model this by learning a normalized attention coefficient for each related neighbor node and then weighted summing all of them. Taking historical interaction relations as an example, the formulas of the $k$-th head for intra-relation aggregation are shown below:

$$\widehat{\boldsymbol{h}}_{n_j}^k = \boldsymbol{W}_{\text{in}}^k \boldsymbol{h}_{n_j} \tag{2}$$

$$c_{n_j}^k = \text{LeakyReLU}\left(\boldsymbol{W}_{\text{intra}}^k{}^{\text{T}}\left[\widehat{\boldsymbol{h}}_{n_i}^k \middle\| \widehat{\boldsymbol{h}}_{n_j}^k\right] p_{n_j}\right) \tag{3}$$

$$p_{n_j} = \text{softmax}(\boldsymbol{W}_{\text{feat}}[t\|w] + \boldsymbol{b}_{\text{feat}}) \tag{4}$$

$$\alpha_{n_j}^k = \frac{\exp\left(c_{n_j}^k\right)}{\sum_{n_{j'} \in \mathcal{N}_{n_i}} \exp\left(c_{n_{j'}}^k\right)} \tag{5}$$

$$\boldsymbol{h}_{n_i}^{\text{his}^k} = \text{softmax}\left(\sum_{n_j \in \mathcal{N}_{n_i}} \alpha_{n_j}^k \widehat{\boldsymbol{h}}_{n_j}^k\right) \tag{6}$$

where $\widehat{\boldsymbol{h}}_{n_j}^k$ is the input of the $k$-th head, $\boldsymbol{W}_{\text{in}}^k$ is the input weight matrix of the $k$-th head and is shared by different relation types. $c_{n_j}^k$ indicates the importance of neighbor node $n_j$ to interacting node $n_i$, which is then normalized across all the choices of $n_j \in \mathcal{N}_{n_i}$ by using the softmax function. $\mathcal{N}_{n_i}$ is the set of related neighbor nodes of $n_i$ in local relation graph $\mathcal{G}_{n_i}$ with historical interaction relations. $^\text{T}$ represents the transposition operation and || denotes the vector concatenation operation. $\boldsymbol{W}_\text{intra}^k$ is the attention weight matrix of the $k$-th head and is specified for each relation type. A fully connected layer is used to learn $p_{n_j}$, which is the weight related to relation attribute $q = (t, w)$ and can be formulated by Equation (4). $\boldsymbol{W}_\text{feat}$ and $\boldsymbol{b}_\text{feat}$ are the parameters of the fully connected layer, which are shared by different relation types and different heads. Once the normalized attention coefficient $\alpha_{n_j}^k$ is obtained for each $n_j \in \mathcal{N}_{n_i}$, a weighted combination of the representations of these nodes is then calculated. Finally, the output $\boldsymbol{h}_{n_i}^{\text{his}\,k} \in \mathbb{R}^d$, which is the neighbor based embedding of $n_i$ based on historical interaction relations of the $k$-th head, is obtained through an activation function.

We execute $K$ independent above processes and then average their outputs, obtaining the following output representation:

$$\boldsymbol{h}_{n_i}^{\text{his}} = \frac{1}{K} \sum_{k=1}^{K} \boldsymbol{h}_{n_i}^{\text{his}\,k} \tag{7}$$

where $\boldsymbol{h}_{n_i}^{\text{his}} \in \mathbb{R}^d$ is the neighbor based embedding of $n_i$ based on historical interaction relations.

*4.2.2 Inter-Relation Aggregation.*

After aggregating the interaction impacts propagated through each relation type, we utilize inter-relation aggregation to combine the neighbor information obtained by all relation types. Since relation types are not equally important in aggregation, we employ the self-attention mechanism [37] to assign different weights to different relation types.

Firstly, we concatenate three embeddings $\boldsymbol{h}_{n_i}^{\text{his}}$, $\boldsymbol{h}_{n_i}^{\text{com}}$, and $\boldsymbol{h}_{n_i}^{\text{seq}}$ as the input of the self-attention mechanism, i.e., $\boldsymbol{H} \in \mathbb{R}^{3 \times d}$, whose query matrix $\boldsymbol{Q}$, key matrix $\boldsymbol{K}$, and value matrix $\boldsymbol{V}$ can be defined as:

$$\boldsymbol{Q} = \boldsymbol{H}\boldsymbol{W}_\text{Q} \tag{8}$$
$$\boldsymbol{K} = \boldsymbol{H}\boldsymbol{W}_\text{K} \tag{9}$$
$$\boldsymbol{V} = \boldsymbol{H}\boldsymbol{W}_\text{V} \tag{10}$$

where $\boldsymbol{W}_\text{Q} \in \mathbb{R}^{d \times d_\text{k}}$, $\boldsymbol{W}_\text{K} \in \mathbb{R}^{d \times d_\text{k}}$, and $\boldsymbol{W}_\text{V} \in \mathbb{R}^{d \times d_\text{v}}$ are learnable parameter matrices.

Then, we aggregate the relation type-specific neighbor based embeddings of $n_i$ using the self-attention function, and the output $\boldsymbol{Z} \in \mathbb{R}^{3 \times d_\text{v}}$ is calculated by:

$$\boldsymbol{Z} = \text{softmax}\left(\frac{\boldsymbol{Q}\boldsymbol{K}^\text{T}}{\sqrt{d_\text{k}}}\right)\boldsymbol{V} \tag{11}$$

where $\sqrt{d_\text{k}}$ is the scaling factor and $d_\text{k} = d_\text{v}$.

At last, an additional fully connected layer is adopted to project the embeddings to a vector space with the desired output dimensions:

$$\boldsymbol{h}'_{n_i} = \text{softmax}(\boldsymbol{W}_\text{out}\boldsymbol{Z} + \boldsymbol{b}_\text{out}) \tag{12}$$

where $\boldsymbol{W}_\text{out}$ and $\boldsymbol{b}_\text{out}$ are the parameter matrices. $\boldsymbol{h}'_{n_i} \in \mathbb{R}^d$ is the neighbor based embedding of $n_i$.

Note that, although the above hierarchical multi-relation aware aggregation only considers the interaction impacts propagated by first-order neighbors, it can be easily extended to consider higher-order neighbors by stacking graph attention layers.

### 4.3 Embedding Update Operation

Intuitively, as interactions occur and time elapses, the properties of users and items will change. In order to model these dynamics, MRATE adopts two RNNs to update the embeddings of interacting users and items, respectively, as shown in Figure 2.

Take interaction $s = (u_i, v_j, t)$ as an example, which represents user $u_i$ interacts with item $v_j$ at time $t$. After this interaction, the user RNN, which is denoted as RNN$_U$, updates the embedding of user $u_i$ after the current interaction, i.e., $u_{i,t_{\text{aft}}}$, by using the embedding of item $v_j$ after the previous interaction, i.e., $v_{j,t_{\text{prev}}}$, as an input. Instead of using the one-hot vector of an item, we adopt the dynamic embedding, which can reflect the current state of the item and obtain more meaningful dynamic user embeddings.

Considering the propagation of interaction impacts and time decay, neighbor information and time interval are also incorporated in our update operation. We utilize the neighbor based embedding of $u_i$, i.e., $h'_{u_i}$, and the time interval between the previous interaction of $u_i$ and the current interaction, i.e., $\Delta_{u_i}$, as the inputs of RNN$_U$, where $h'_{u_i}$ is obtained by using the aforementioned hierarchical multi-relation aware aggregation.

Similarly, RNN$_V$ updates the embedding of item $v_j$ after the current interaction, i.e., $v_{j,t_{\text{aft}}}$, by using the embedding of user $u_i$ after the previous interaction, i.e., $u_{i,t_{\text{prev}}}$, the neighbor based embedding of $v_j$, i.e., $h'_{v_j}$, and the time interval between the previous interaction of $v_j$ and the current interaction, i.e., $\Delta_{v_j}$, as inputs. Formally, the update operations are given by:

$$u_{i,t_{\text{aft}}} = \text{sigmoid}\left(W_1^{\text{u}} u_{i,t_{\text{prev}}} + W_2^{\text{u}} v_{j,t_{\text{prev}}} + W_3^{\text{u}} h'_{u_i} + W_4^{\text{u}} t_{u_i}\right) \tag{13}$$

$$v_{j,t_{\text{aft}}} = \text{sigmoid}\left(W_1^{\text{v}} v_{j,t_{\text{prev}}} + W_2^{\text{v}} u_{i,t_{\text{prev}}} + W_3^{\text{v}} h'_{v_j} + W_4^{\text{v}} t_{v_j}\right) \tag{14}$$

where $W_1^{\text{u}}$, $W_2^{\text{u}}$, $W_3^{\text{u}}$, and $W_4^{\text{u}}$ are learnable parameters of RNN$_U$. $W_1^{\text{v}}$, $W_2^{\text{v}}$, $W_3^{\text{v}}$, and $W_4^{\text{v}}$ are learnable parameters of RNN$_V$. $t_{u_i}$ and $t_{v_j}$ are the representations of time intervals $\Delta_{u_i}$ and $\Delta_{v_j}$, which are obtained by a shared fully connected layer, respectively. Note that, RNN$_U$ is shared by all users to update user embeddings and RNN$_V$ is shared by all items to update item embeddings. The hidden states of RNN$_U$ and RNN$_V$ represent the user and item embeddings, respectively.

### 4.4 Training

#### 4.4.1 Objective Function.

Let user $u_i$ currently interact with item $v_j$. We use the task of predicting which item $u_i$ will interact with right before the current interaction time to train MRATE.

A fully connected layer is employed to calculate the predicted embedding of item $v_j$, i.e., $\tilde{v}_{j,t_{\text{bef}}}$, based on the embedding of $u_i$ after the previous interaction, i.e., $u_{i,t_{\text{prev}}}$, and the neighbor based embedding of $u_i$, i.e., $h'_{u_i}$, which can be formulated by:

$$\tilde{v}_{j,t_{\text{bef}}} = \text{softmax}\left(W_1 u_{i,t_{\text{prev}}} + W_2 h'_{u_i} + b\right) \tag{15}$$

where $W_1$, $W_2$, and $b$ are the parameter matrices of the fully connected layer.

For simplicity, take the embedding of $v_j$ after the previous interaction, i.e., $v_{j,t_{\text{prev}}}$, as the ground truth embedding. The prediction loss of MRATE is defined as the mean squared error (MSE) between the predicted embedding and the ground truth embedding, which can be calculated by:

$$\mathcal{L} = \sum_{(u_i,v_j,t)\in\mathcal{S}} \left\|\tilde{v}_{j,t_{\text{bef}}} - v_{j,t_{\text{prev}}}\right\|_2 + \lambda_U \left\|u_{i,t_{\text{aft}}} - u_{i,t_{\text{prev}}}\right\|_2 + \lambda_I \left\|v_{j,t_{\text{aft}}} - v_{j,t_{\text{prev}}}\right\|_2 \tag{16}$$

## ALGORITHM 1: T-n-Batch algorithm

**Input**: Temporal interaction network $S = \{s_i\}_{i=1}^{N}$
**Output**: Batch sequence $B: B_k = \{s_{k,1}, ..., s_{k,m}\}$
Initialize all users' last-batch index $lastU[u] \leftarrow 0, \forall u \in \mathcal{U}$
Initialize all items' last-batch index $lastV[v] \leftarrow 0, \forall v \in \mathcal{V}$
Initialize batches $B_k \leftarrow \emptyset, \forall k \in [1, N]$
Initialize non-empty batch count $C \leftarrow 0$
**For** $i = 1, 2, ..., N$ **do**:
   /*Taking interaction $(u_i, v_j, t)$ as an example */
   Extract neighbor node set $\mathcal{N} \leftarrow \mathcal{N}_{u_i} \cup \mathcal{N}_{v_j}$
   Find $idxN$ among the last-batch indexes of all nodes in $\mathcal{N}$
   Find the $i$-th interaction's batch id $idx \leftarrow \max(lastU[u_i], lastV[v_j], idxN) + 1$
   Add the $i$-th interaction to correct batch $B_{idx} \leftarrow B_{idx} \cup \{(u_i, v_j, t)\}$
   Update user $u_i$'s last-batch index $lastU[u_i] \leftarrow idx$
   Update item $v_i$'s last-batch index $lastV[v_j] \leftarrow idx$
   Update batch count $C \leftarrow max(C, idx)$
**End for**
**Return** $\{B_1, ..., B_C\}$

where the first loss term is the predicted loss, and the last two terms are regularized losses, which are used to avoid the dynamic user and item embeddings vary too much, respectively. $\lambda_U$ and $\lambda_I$ are scaling parameters and $\| \ \|_2$ denotes L2 distance. Note that, since MRATE directly outputs the predicted embeddings of items, it is not necessary to use negative sampling during training.

### 4.4.2 Batching Algorithm.

Since the order of interactions in temporal interaction networks is important, the temporal dependencies between interactions need to be maintained during training, i.e., interaction $s_i$ is processed before $s_j$ if $i < j$. In addition, it is necessary to parallelize the training of MRATE, as a large number of interactions will lead to the slow training process. Thus, it is challenging to create the training batches because of two requirements [22]: (1) Interactions in a same batch should be processed in parallel; (2) Processing all batches in the increasing order of their indexes should maintain the time dependencies. Here, we propose a batching algorithm called t-n-Batch to meet the above two requirements, which extent t-batch [22] to take neighbor nodes into consideration.

The detailed batch splitting process is shown in Algorithm 1. Given temporal interaction network $S = \{s_i\}_{i=1}^{N}$, we first initialize $N$ empty batches, and then add each interaction to the corresponding batch in the temporally order of interactions. Taking interaction $(u_i, v_j, t)$ as an example, $lastU[u_i]$ and $lastV[v_j]$ are last-batch indexes, which denote the maximum indexes of batches that have an interaction involving user $u_i$ and item $v_j$, respectively. $idxN$ is the maximum one among the last-batch indexes of all neighbor nodes in local relation graphs of $u_i$ and $v_j$, which are constructed based on all historical interactions in advance. Note that, using all neighbor nodes will get the same embeddings as using related neighbor nodes. Considering the above two requirements, each node can only appear at most once in a batch, and the $i$-th and $i + 1$-th interactions of each

---
**ALGORITHM 2**: The training process of MRATE
---
**Input**: Temporal interaction network $\mathcal{S} = \{s_i\}_{i=1}^{N}$, epoch number $R$
**Output**: User and item embeddings after all interactions, fine-tuned MRATE
Initialize MRATE with random network parameters
Get batches by the t-n-Batch algorithm
**For** $ep = 1,2,\ldots,R$ **do**:
  Initialize previous user embedding $\boldsymbol{u}_{t_{\text{prev}}}, \forall u \in \mathcal{U}$
  Initialize previous item embedding $\boldsymbol{v}_{t_{\text{prev}}}, \forall v \in \mathcal{V}$
  **For** $i = 1,2,\ldots, batch\ number$ **do**:
    $loss \leftarrow 0$
    **For** each interaction in the $i$-th batch **do in parallel**:
      /*Taking interaction $(u_i, v_j, t)$ as an example    */
      Calculate neighbor based user embedding $\boldsymbol{h}'_{u_i}$
      Calculate neighbor based item embedding $\boldsymbol{h}'_{v_j}$
      Predict item embedding $\widetilde{\boldsymbol{v}}_{j,t_{\text{bef}}}$ using Equation (15)
      Get ground truth embedding $\boldsymbol{v}_{j,t_{\text{prev}}}$
      Calculate prediction loss $loss \leftarrow loss + \left\| \widetilde{\boldsymbol{v}}_{j,t_{\text{bef}}} - \boldsymbol{v}_{j,t_{\text{prev}}} \right\|_2$
      Update user embedding after the current interaction $\boldsymbol{u}_{i,t_{\text{aft}}} \leftarrow \text{RNN}_U(\boldsymbol{u}_{i,t_{\text{prev}}}, \boldsymbol{v}_{j,t_{\text{prev}}}, \boldsymbol{h}'_{u_i}, \Delta_{u_i})$
      Update item embedding after the current interaction $\boldsymbol{v}_{j,t_{\text{aft}}} \leftarrow \text{RNN}_V(\boldsymbol{v}_{j,t_{\text{prev}}}, \boldsymbol{u}_{i,t_{\text{prev}}}, \boldsymbol{h}'_{v_j}, \Delta_{v_j})$
      Add user and item embedding drifts to loss $loss \leftarrow loss + \lambda_U \left\| \boldsymbol{u}_{i,t_{\text{aft}}} - \boldsymbol{u}_{i,t_{\text{prev}}} \right\|_2 + \lambda_I \left\| \boldsymbol{v}_{j,t_{\text{aft}}} - \boldsymbol{v}_{j,t_{\text{prev}}} \right\|_2$
      Update previous user embedding $\boldsymbol{u}_{i,t_{\text{prev}}} \leftarrow \boldsymbol{u}_{i,t_{\text{aft}}}$
      Update previous item embedding $\boldsymbol{v}_{j,t_{\text{prev}}} \leftarrow \boldsymbol{v}_{j,t_{\text{aft}}}$
    **End for**
    Back-propagate loss and update model parameters
  **End for**
**End for**
**Return** $\boldsymbol{u}_{t_{\text{aft}}} \forall u \in \mathcal{U}, \boldsymbol{v}_{t_{\text{aft}}} \forall v \in \mathcal{V}$, fine-tuned MRATE
---

node need to be assigned to batches $B_k$ and $B_l$, where $k < l$. Therefore, the interaction $(u_i, v_j, t)$ is assigned to the batch with index $\max(lastU[u_i], lastV[v_j], idxN) + 1$.

### 4.4.3 Training Algorithm.

The detailed training process for MRATE is shown in Algorithm 2. Based on the t-n-Batch algorithm, the batches that meet the aforementioned requirements are created. Since the outputs of hierarchical multi-relation aware aggregation are a part of the inputs of the paired RNNs, all network parameters are updated by the loss defined in Equation (16) during training.

## 5 EXPERIMENTS

In this section, we present the experimental datasets, settings, and results to evaluate the proposed method MRATE. The task we used here is predicting which item a user will interact with given all previous interactions. Specifically, given a query with current time $t$ and user $u_i$, we first obtain the predicted embedding $\tilde{v}_{t_{\text{bef}}}$ of the next interact item using Equation (15). Then, we calculate the L2 distances between $\tilde{v}_{t_{\text{bef}}}$ and the true embeddings of all items, and finally recommend top $K$ items with small L2 distances to $u_i$.

### 5.1 Datasets

MRATE is evaluated on three public datasets, i.e., Reddit[1], LastFM [38], and Wikipedia[2], all of which consist of a large number of interactions and are widely used for temporal interaction network embedding.

Reddit consists of one month of posts made by users on subreddits. We select 1,000 most active subreddits as items and 10,000 most active users from this dataset to avoid the influence of noise [22], resulting in 672,447 interactions.

LastFM consists of one month of listening records made by users on the social music platform Last.fm. We select 1,000 most listened songs as items and 1,000 most active users [38], resulting in 1,293,103 interactions.

Wikipedia consists of one month of edits made by users on Wikipedia pages. We select 1,000 most edited pages as items and editors who made at least 5 edits as users [22], resulting in 8,227 users and 157,474 interactions.

### 5.2 Experimental Settings

MRATE is implemented based on PyTorch software library, and the training and testing are conducted on GPUs. The code is released on GitHub[3]. Adam algorithm [39] is used to optimize model parameters, and the initial learning rate is set to 0.001. Through parameter tuning, the dimension of the sequence based embedding is set to 100, the length of sliding window of Doc2Vec is set to 5, the threshold $\mu$ is set to 0.5, the number of attention heads $K$ is set to 3, and the length of sliding window of RNN is set to 1. Other arguments are set by default.

In order to simulate the real situation, all temporal interaction network embedding models are trained by splitting the data by interaction time. Thus, we train all models on the first 80% interactions, validate on the next 10% interactions, and test on the remaining 10% interactions. Note that, the validating and testing processes of MRATE also need to update model parameters. Different from the training process, the parameters are updated whenever an interaction is processed. For a fair comparison, all models in the following experiments are run for 50 epochs and the dimension of dynamic user and item embeddings is set to 120.

Mean reciprocal rank (MRR) and recall@10 are adopted to measure the performance of a temporal interaction network embedding model. MRR is the average of the reciprocal ranks, which can be formulated by Equation (17), and recall@10 is the percentage of the ground truth items ranked in the top 10.

$$\text{MRR} = \frac{1}{M}\sum_{i=1}^{M}\frac{1}{rank_i} \quad (17)$$

where $M$ denotes the number of samples in the test set, and $rank$ denotes the rank of the ground truth item.

---

[1] http://files.pushshift.io/reddit
[2] https://meta.wikimedia.org/wiki/Data_dumps
[3] https://github.com/ShansYu/mrate

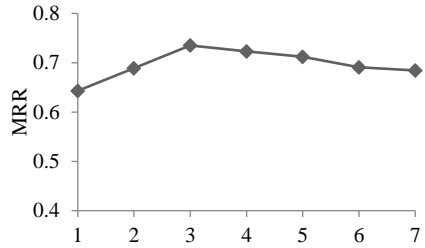

Figure 4: The impact of parameter $T$.

### 5.3 Parameter Evaluation

There is a key parameter when mining common interaction relations, i.e., time slot $T$. Here, we investigate how $T$ affects the performance of MRATE. We increase the value of $T$ from 1 day to 7 days with an incremental step of 1 day and record the performance on the validation set.

Figure 4 illustrates the MRRs on different values of $T$ on Reddit dataset. The performance of MRATE first increases along with the increasing value of $T$ and then decreases. It suggests that a larger value of $T$ can maintain more effective information for mining common interaction relations. But when $T$ is too large, it would introduce noise and show worse performance. Thus, we set $T$ to 3 days, which is fixed in the following experiments.

### 5.4 Ablation Experiments

MRATE mines the relations of three types between nodes from historical interactions. To investigate the effectiveness of each relation type and the components of MRATE, we compare MRATE with the following variants:

**MRATE-w/o-his**: MRATE-w/o-his only uses common interaction relations and interaction sequence similarity relations to construct local relation graphs, and the rest is consistent with MRATE.

**MRATE-w/o-com**: MRATE-w/o-com is similar with MRATE-w/o-his, but only uses historical interaction relations and interaction sequence similarity relations.

**MRATE-w/o-seq**: MRATE-w/o-seq is similar with MRATE-w/o-his, but only uses historical interaction relations and common interaction relations.

**MRATE-w/o-att**: MRATE-w/o-att removes hierarchical multi-relation aware aggregation from MRATE, which assigns the same weights to all neighbor nodes for neighbor information aggregation. Its other components are the same as MRATE.

**MRATE-w/o-pro**: MRATE-w/o-pro only uses paired RNNs to update the embeddings of interacting users and items, which ignores neighbor information.

For fairness, all variants are well-tuned and follow the same experimental settings as MRATE. The results of these methods are shown in Table 2, from which we can discern the following tendencies:

1) MRATE outperforms all variants on three datasets, which indicates that the relations of multiple types, hierarchical multi-relation aware aggregation, and neighbor information used in MRATE can improve the performance of temporal interaction network embedding.

Table 2: The performance of MRATE and variants

|  | Reddit | | LastFM | | Wikipedia | |
| --- | --- | --- | --- | --- | --- | --- |
|  | MRR | Recall@10 | MRR | Recall@10 | MRR | Recall@10 |
| MRATE-w/o-his | 0.672 | 0.749 | 0.172 | 0.264 | 0.692 | 0.735 |
| MRATE-w/o-com | 0.617 | 0.728 | 0.169 | 0.258 | 0.638 | 0.674 |
| MRATE-w/o-seq | 0.695 | 0.803 | 0.183 | 0.279 | 0.718 | 0.782 |
| MRATE-w/o-att | 0.648 | 0.782 | 0.159 | 0.243 | 0.621 | 0.753 |
| MRATE-w/o-pro | 0.542 | 0.688 | 0.125 | 0.226 | 0.539 | 0.596 |
| MRATE | **0.723** | **0.821** | **0.189** | **0.297** | **0.748** | **0.792** |

2) The performance of MRATE-w/o-pro and MRATE-w/o-att is obviously worse than MRATE, which demonstrates the effectiveness of considering neighbor information. In addition, using hierarchical multi-relation aware aggregation can capture the significances of different neighbor nodes and achieve better performance.

3) The impacts of three relation types can be ranked as follows: common interaction relations > historical interaction relations > interaction sequence similarity relations. The reason might be that common interaction relations model the interaction behaviors of users and items, which directly reflects the similarity of two nodes' properties. Interaction sequence similarity relations have the least impact on temporal interaction network embedding. It might be because all interactions of two nodes are used to calculate the interaction sequence similarity between them, which cannot capture the short-term similarity and is not accurate enough.

## 5.5 Comparison with the State-of-the-Art Methods

To show the competitive performance of MRATE, we compare it with the state-of-the-art methods.

**Time-LSTM**: Time-LSTM [16] equips a LSTM with time gates to model time intervals in users' interaction sequences. Since Time-LSTM-3 cell performs best in the original paper, we use it to obtain the embeddings of users and items. Note that, we use the one-hot vectors of items as inputs.

**RRN**: RRN [17] adopts two LSTMs to model the dynamics of users and items, and the inputs of them are the one-hot vectors of items and users, respectively.

**CDTNE**: CDTNE [20] first constructs a graph based on historical interactions, in which a node denotes a user or an item, an edge indicates an interaction between two connected nodes, and the edge attribute is the interaction time. Random walks that obey interaction time are utilized to obtain node sequences and then node embeddings are obtained by the skip-gram model.

**JODIE**: JODIE [22] employs paired RNNs to update the embeddings of the interacting user and item when an interaction occurs. In addition to using the interaction data, this method also takes interaction features into consideration, e.g., the post text and the edit text. To ensure fairness, we use this method as a baseline without considering interaction features.

**DGNN**: DGNN [25] takes neighbor information into account. In addition to updating the embeddings of interacting nodes by two RNNs, this method also uses other two RNNs to update the embeddings of neighbor nodes, which have historical interaction relations with interacting nodes.

For fairness, all the above methods follow the same experimental settings as MRATE and the parameters of these methods are optimized.

The performance of these methods is shown in Table 3, and the following tendencies can be discerned:

Table 3: The performance of MRATE and the state-of-the-art methods

|  | Reddit | | LastFM | | Wikipedia | |
| --- | --- | --- | --- | --- | --- | --- |
|  | MRR | Recall@10 | MRR | Recall@10 | MRR | Recall@10 |
| Time-LSTM | 0.375 | 0.553 | 0.062 | 0.129 | 0.235 | 0.328 |
| RRN | 0.453 | 0.607 | 0.086 | 0.178 | 0.412 | 0.587 |
| CTDNE | 0.165 | 0.254 | 0.015 | 0.027 | 0.039 | 0.072 |
| JODIE | 0.542 | 0.688 | 0.125 | 0.226 | 0.566 | 0.632 |
| DGNN | 0.614 | 0.732 | 0.178 | 0.267 | 0.725 | 0.732 |
| MRATE | **0.723** | **0.821** | **0.189** | **0.297** | **0.748** | **0.792** |

1) On three datasets, MRATE outperforms all compared methods. It demonstrates that constructing local relation graphs by defining three relation types can discover implicit relations between nodes and employing hierarchical multi-relation aware aggregation can fully utilize the relations.

2) Time-LSTM and RRN both perform worse than JODIE. The reason might be that using the dynamic embeddings instead of the static embeddings of items as the inputs of RNNs or LSTMs to update user embeddings can capture the current states of items and obtain more effective user embeddings. In addition, RRN has better performance than Time-LSTM. It might be because RRN updates not only user embeddings but also item embeddings by two LSTMs.

3) CTDNE performs the worst among all methods. The reason might be that CTDNE constructs a static graph and cannot effectively model the dynamics of users and items.

4) DGNN performs better than JODIE. The reason might be that DGNN takes neighbor information into consideration. Four RNNs are used to update the embeddings of interacting nodes and neighbor nodes, which can model the propagation of interaction impacts and obtain effective user and item embeddings.

## 6 CONCLUSIONS AND FUTURE WORK

In this paper, a multi-relation aware temporal interaction network embedding method (MRATE) is proposed. When an interaction occurs, MRATE first mines the relations of three types, i.e., historical interaction relations, common interaction relations, and interaction sequence similarity relations, to construct the local relation graphs of interacting nodes. Then, hierarchical multi-relation aware aggregation is introduced to obtain the neighbor based embeddings of interacting nodes. Finally, two RNNs are utilized to update the embeddings of the interacting user and item. We evaluate MRATE on three public datasets and the experimental results justify the advantage of this method.

In the future, we will extend our work in the following directions. First, we will mine more implicit relation types to obtain richer information from historical interactions. Second, we will introduce edge graphs to model the relations between edges and make full use of them.

### ACKNOWLEDGEMENTS

This work is supported by the National Key Research and Development Program of China (No. 2018YFB0505000) and the Fundamental Research Funds for the Central Universities (No. 2020QNA5017).